\PassOptionsToPackage{table}{xcolor}

\documentclass[sigconf,authorversion,nonacm]{acmart}

\usepackage{newtxmath}
\usepackage{xcolor}

\usepackage{multirow}
\usepackage{tcolorbox}
\AtBeginDocument{%
  \providecommand\BibTeX{{%
    \normalfont B\kern-0.5em{\scshape i\kern-0.25em b}\kern-0.8em\TeX}}}
\author{Chang Xiao}
\affiliation{%
   \institution{Adobe Research}
    \country{San Jose, California, USA}
 }

 \author{Brenda Z. Yang}
\affiliation{%
   \institution{Columbia University}
    \country{New York, New York, USA} 
 }

\newcommand{\figref}[1]{Figure~\ref{fig:#1}}
\newcommand{\tabref}[1]{Table~\ref{tab:#1}}

\makeatletter
\renewcommand{\paragraph}{%
  \@startsection{paragraph}{4}%
  {\z@}{0.60ex \@plus 1ex \@minus .15ex}{-1em}%
  {\normalfont\normalsize\bfseries}%
}
\makeatother

\begin{document}

\title{LLMs May Not Be Human-Level Players, But They Can Be Testers: Measuring Game Difficulty with LLM Agents}

\renewcommand{\shortauthors}{Anonymous}

\begin{abstract}
Recent advances in Large Language Models (LLMs) have demonstrated their potential as autonomous agents across various tasks. One emerging application is the use of LLMs in playing games. In this work, we explore a practical problem for the gaming industry: \textbf{Can LLMs be used to measure game difficulty?} We propose a general game-testing framework using LLM agents and test it on two widely played strategy games: \textit{Wordle} and \textit{Slay the Spire}. Our results reveal an interesting finding: although LLMs may not perform as well as the average human player, their performance, when guided by simple, generic prompting techniques, shows a statistically significant and strong correlation with difficulty indicated by human players. This suggests that LLMs could serve as effective agents for measuring game difficulty during the development process. Based on our experiments, we also outline general principles and guidelines for incorporating LLMs into the game testing process.

\end{abstract}



\maketitle

\section{Introduction}

Since the dawn of the digital era, video games have become a significant part of human culture. From early, primitive games like \textit{Pong} to modern, high-profile titles like \textit{Grand Theft Auto}, this medium has evolved into a multibillion-dollar industry, playing a pivotal role in the entertainment sector. According to statistics from the \textit{Steam} gaming platform~\cite{statista_steam_games}, more than ten thousand new video games are released each year.

A key aspect of developing a successful video game lies in achieving the right balance of difficulty. Players seek challenges that feel rewarding but not overwhelming, and poorly balanced difficulty can lead to frustration and disengagement~\cite{juul2013art}. In game design theory~\cite{rollings2003andrew}, this is referred to as designing a smooth difficulty curve, where players face increasing challenges that remain manageable as they progress through different levels.

A notable example illustrating the importance of balanced difficulty is the recent hit \textit{Elden Ring: Shadow of the Erdtree}. Upon its initial release, the game was criticized for being excessively difficult~\cite{tassi_eldenring_2024, pcgamer2024erdtree}, with only a limited number of players able to progress through even the first few areas. In response to widespread complaints, the developers had to release a patch to reduce the difficulty, leading to additional development costs and reduced consumer satisfaction.

Ensuring that a game's difficulty curve aligns with the designer’s vision requires rigorous testing throughout the development phase. Traditionally, human players are recruited to assess the game's difficulty~\cite{aponte2011measuring}. These testers comprehensively navigate through various levels and scenarios to plot out the difficulty curve and see if it match with the developer's intent. While this method is thorough, it is also inherently tedious, complex, and resource-intensive, requiring significant time and human power, especially when iterative changes are necessary.

To address this issue, researchers and developers have explored automated testing techniques, using AI to simulate human players and analyze AI's performance to assess game difficulty~\cite{guerrero2018using, perez2016general}. However, there is still uncertainty regarding how accurately AI can approximate the abilities of real players. Additionally, training such AI models, particularly those utilizing advanced techniques like deep reinforcement learning, demands significant computational resources. Moreover, these models are often tailored to specific games, limiting their reusability across different games. This raises concerns about the cost-effectiveness of creating game-specific AI systems for game testing~\cite{shao2019survey}.

As a result, there is a growing need for creating a General Video Game Playing (GVGP) agent that can be easily deployed across different games while exhibiting human-like behavior to a reasonable extent. The recent rise of Large Language Models (LLMs) has sparked interest among researchers and developers in their potential to act as GVGP agents. Research has demonstrated that LLMs can play video games effectively, and with carefully designed system architectures and prompts, they can even perform at an advanced player level in specific cases~\cite{hu2024pok, huang2024pokergpt, wangvoyager}.

Given the reasoning capabilities of LLMs and their demonstrated success in gameplay scenarios, we aim to explore whether LLMs can be effectively utilized in game testing, specifically to assess game difficulty. Our target is to establish a general framework for LLMs to play games without extra fine-tuning, and assess difficulty in a way that closely aligns with human. 

Our contributions to this work are threefold:

\begin{itemize}
    \item We propose a general game-testing framework that leverages LLMs to evaluate whether the various game challenges reflect the difficulty intended by the game developers. 
    \item We deployed this framework in two popular video games---\textit{Wordle} and \textit{Slay the Spire}. Our findings show that LLMs exhibit a strong correlation with human on difficulty across various challenges, even though they generally do not match the gameplay performance (e.g., scores, win rates) of the average human player. Specifically, when the LLM performed poorly on certain challenges, human players also found them difficult, and vice versa. This suggests that LLMs can serve as effective testers for assessing relative difficulty of different challenges.
    \item Based on our experiments using LLM agents to play games, we outline several guidelines for effectively leveraging LLMs to evaluate game difficulty. These guidelines could offer insights for the gaming industry and help the development of LLM-based game-testing environments.
\end{itemize}

To the best of our knowledge, our work is the first to explore the feasibility of using LLMs to measure game difficulty, with validation based on human gameplay data.

\subsection{Measurement of Difficulty}

To this point, we have used the term ``difficulty'' without providing a formal definition. According to prior literature~\cite{aponte2009scaling, aponte2011measuring, juul2003game}, difficulty, in the context of video game, is generally related to the player's skills or abilities to overcome a given ``challenge''. A ``challenge'' in a game can be understood as a sub-game: a rule-based system with variable and quantifiable outcomes~\cite{juul2003game}.

While difficulty for a specific challenge could theoretically be represented by an absolute metric, in game development, what is often more important is the relative scaling of difficulty across different challenges~\cite{aponte2009scaling}. A well-structured difficulty curve helps players achieve a state of \textit{flow}~\cite{czikszentmihalyi1990flow}, where the task is neither too hard nor too easy, thereby increasing player engagement and enjoyment.
When applying LLMs for difficulty assessment, the goal should not be to obtain an absolute difficulty value for any individual challenge. Instead, the focus should be on determining the relative difficulty and its scaling among a series of challenges. 

As discussed by Aponte et al.~\cite{aponte2011measuring}, difficulty can be measured experimentally by having many players attempt a specific set of challenges and evaluating the quality of their performance. The quality measurement can involve any game-related statistics that correlate with difficulty, such as the player's win rate in overcoming a challenge, the score achieved, or the time taken. The specific metric chosen depends on the context and game, as there is no universally correct measure of difficulty.


\section{Related Work}
\subsection{AI for Game Testing}
Before the rise of intelligence models in this decade, the term ``AI'' was commonly used in the context of gaming, where it referred to algorithms used to control non-player characters or automate in-game mechanics. At the same time, due to the high cost of human testing in game development, the idea of using AI for game testing has been explored since the early days of video games. Work by Macleod~\cite{macleod2005game} investigated this by simulating gameplay in \textit{Perudo}, a bidding dice game, using a multi-agent system. Similarly, Kirby et al.~\cite{kirby2011introduction} studied the classic game \textit{Minesweeper} by replacing human players with a rule-based algorithm, finding that a simple algorithm could solve most cases. Although these are relatively simple games with straightforward mechanics, the studies demonstrated the potential of using rule-based AI for game testing. Nowadays, the concept of automated game testing has been widely studied in game research~\cite{aponte2011measuring, guerrero2018using} and integrated into modern game development frameworks such as Unity~\cite{unity}.




As video games have evolved, their scale has expanded, and the mechanics have become increasingly complex~\cite{aponte2009scaling}. In modern video game development, creating AI strategies for these games has become as challenging as, if not more challenging than, designing the games themselves~\cite{perez2016general}. As a result, researchers are exploring more advanced and generalized methods for developing game AI, with a focus on potentially creating GVGP agents. Notably, the use of Deep Learning (DL) to train game AI has seen tremendous success in recent years, whether through supervised learning from human gameplay~\cite{swiechowski2018improving, barriga2019improving, ye2020supervised} or reinforcement learning via self-play~\cite{risi2020chess, heinrich2016deep, silver2018general}. These approaches have been applied to a wide range of popular video games, including  \textit{Atari}~\cite{mnih2013playing}, \textit{Starcraft}~\cite{vinyals2019grandmaster, ontanon2013survey}, and \textit{Doom}~\cite{lample2017playing}. These techniques have enabled game AI to reach superhuman levels of performance, as measured by metrics like scores and win rates. For a comprehensive review of DL-based game AI, refer to the survey by Shao et al.~\cite{shao2019survey}. However, many of these works have focused on building better AI models rather than on utilizing AI to assist in the game development process.

To address this gap, researchers have recently expanded their focus beyond simply developing stronger game AI, to explore how AI can provide insights into game design and player experience. For example, Zhu et al.~\cite{zhu2021player} analyzed over 20 games, identifying dominant player-AI interaction metaphors and patterns in these games. Villareale et al.~\cite{villareale2022want} conducted a study where participants played against adversarial AI models, investigating how players developed mental models of the AI during gameplay. Liang et al.~\cite{liang2019implicit} offered insights into how AI can enhance the natural and efficient communication of actionable information in games involving human-AI collaboration. These studies suggest that AI systems can offer valuable insights for game design. Additionally, there is a growing body of research discussing the design space for AI's role in games and how it can improve player experience~\cite{treanor2015ai, zhu2021player, guzdial2019friend}. 

Despite this progress, creating AI systems for games remains a significant challenge. Game AI often needs to be designed or trained on a case-by-case basis for each game, particularly for DL-based AI, which demands substantial computational resources. Another major challenge is the lack of interpretability in DL-based AI systems~\cite{shao2019survey}. For instance, it can be difficult to understand why an AI chooses a particular move over another, which limits the insights that game developers can derive from these systems. Therefore, it is still unclear how AI can be used to increase the productivity of game development.

\subsection{LLM Agent for Gaming}
The increasing capabilities of LLMs have demonstrated their potential across a wide range of tasks, from summarizing an article to complex code generation, leading people to consider their potential as universal agents for GVGP. This potential is attributed to their ability to process diverse textual inputs, interpret natural language instructions, and reason effectively to generate appropriate outputs. These features make LLMs well-suited for processing game-related data and producing human-like actions with minimal training effort.

Similar to the trend in DL-based AI, there is a significant amount of work on developing LLM agents to achieve better gameplay performance. Examples of these efforts span various game genres, including conversational games like \textit{Werewolf}~\cite{xu2023language} and \textit{Avalon}~\cite{light2023avalonbench}, board and card games like \textit{Poker}~\cite{huang2024pokergpt}, \textit{Chess}~\cite{feng2024chessgpt}, and \textit{Crossword}~\cite{saha2024language}, as well as more intricate video games like \textit{Street Fighter}~\cite{LLMColosseum2024}, \textit{StarCraft II}~\cite{ma2023large}, \textit{Pok\'emon}~\cite{hu2024pok}, \textit{Civilization}~\cite{qi2024civrealm}, \textit{Minecraft}~\cite{wangvoyager} and \textit{Slay the Spire}~\cite{bateni2024sts}. Many of these studies suggest that with appropriate prompting techniques and system structures, LLMs can exhibit human-like behaviors in gameplay and outperform traditional heuristic-based AI algorithms. Recent surveys have summarized this line of work on using LLMs for playing games~\cite{hu2024survey, zhang2024llm}. However, most of these studies primarily focus on developing better LLM agents for optimal gameplay performance or as AI adversaries to human player, with little attention given to how LLMs could be utilized in other parts of game development.

Therefore, a growing body of recent research explores additional possibilities for using LLMs in different aspects of game development. For example, besides being a player, LLMs could also serve as non-player characters (NPCs) or commentators~\cite{shanahan2023role, ranella2023towards}, enriching the player's experience. Furthermore, LLMs' abilities in text summarization and generating responses to open-ended input make them ideal for acting as Game Masters (GMs) in classical tabletop role-playing games like \textit{Dungeons \& Dragons}~\cite{tychsen2005game}. One of the first notable text-based adventures managed by a fine-tuned version of GPT-2 was \textit{AI Dungeon}~\cite{treanor2015ai}. Several works have also explored how LLMs can assist human GMs during game sessions~\cite{zhu2023calypso, kelly2023towards, acharya2023shoelace}.

In addition to these roles, LLMs' generative abilities are crucial for Procedural Content Generation (PCG) in games. By structuring game content as text, LLMs can generate new level designs for various games, such as \textit{Sokoban}~\cite{todd2023level} and \textit{Super Mario}~\cite{sudhakaran2024mariogpt}. Given their capacity to incorporate world knowledge and generate content, LLMs are anticipated to play a significant role in future game design. Gallotta et al.~\cite{gallotta2024large} outlined the potential contributions of LLMs to gameplay and development and discussed a roadmap for integrating LLMs into the game design process.

Despite these diverse contributions, an underexplored landscape remains in using LLM agents to assess game difficulty. We aim to explore their potential and examine how closely LLMs can simulate human in evaluating game difficulty.

\subsection{LLM as Human-like Simulator}
Outside the gaming domain, a considerable body of research has explored leveraging LLMs as human simulators across various fields, including economics~\cite{li2024econagent, horton2023economics, aher2023eco}, politics~\cite{wu2023politics, qi2024representation}, healthcare~\cite{peng2023medical, tustumi2023future}, and social science~\cite{ziems2024social, park2023generative, manning2024automated}. These studies suggest that, to a certain extent, LLMs can effectively model human behavior, decision-making, and social interactions across different contexts. Given these promising applications, we are interested in understanding whether similar capabilities can be observed in the gaming domain.

\section{Framework Overview and Game Choice}

\begin{figure*}[t!]
    \centering
    \includegraphics[width=0.98\textwidth]{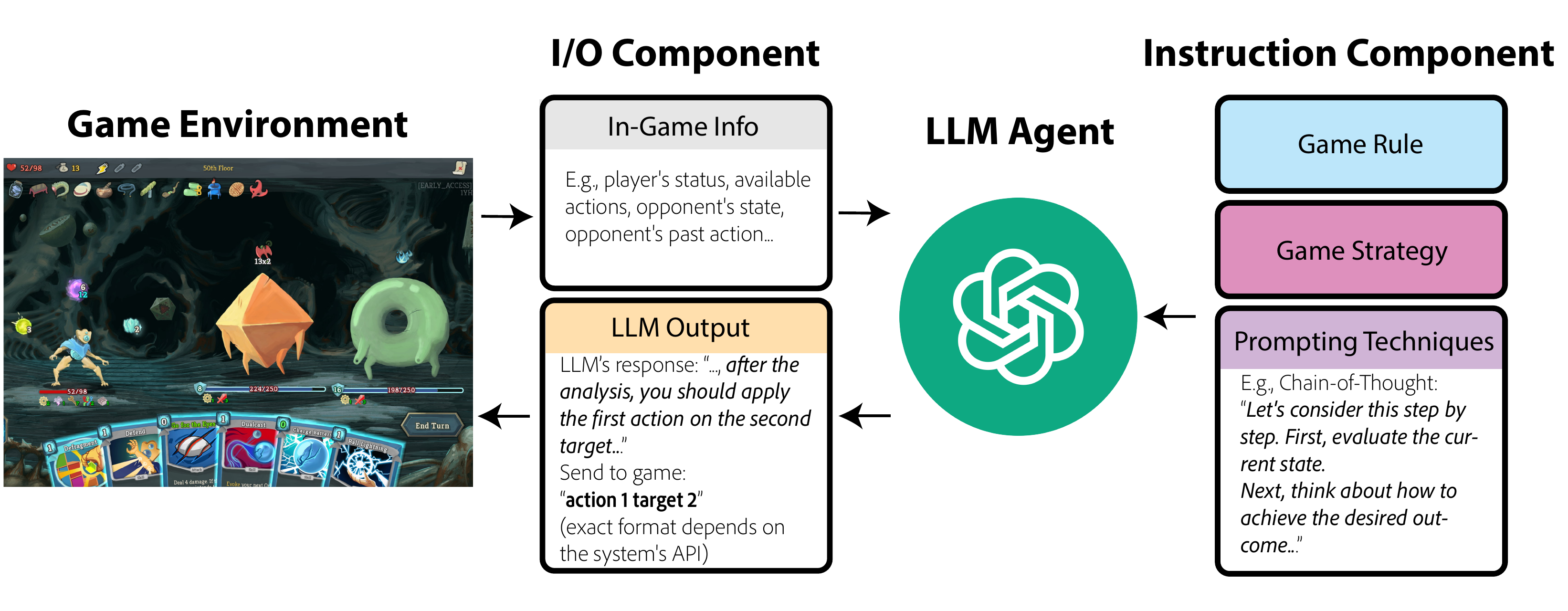}
    \caption{Overview of the proposed framework for LLM-based game difficulty testing. In each step of the game loop, game information is extracted via APIs, converted into natural language, and processed by the LLM along with additional details, such as game rules and strategies, using prompting techniques like Chain-of-Thought. The LLM outputs a suggested player action, which is translated into API calls or keyboard/mouse events to execute in-game. The loop continues until the challenge is completed or failed.}
    \label{fig:framework}
\end{figure*}

Our goal is to provide a general framework for game difficulty measurement, using the LLM's gameplay performance as the difficulty metric.  \figref{framework} depicts our framework architecture. Our framework includes two major components: the game I/O component and the instruction component.

\begin{itemize}
    \item \textbf{Game I/O Component}: This component handles the interaction between the LLM and the game environment. It parses the game state, summarizes the necessary information for the LLM, and converts the LLM's responses into executable game actions.
    \item \textbf{Instruction Component}: This component consists of three parts: Game Rules, Game Strategies, and Prompting Techniques. The Game Rules provide essential explanations of how the game operates, ensuring that LLMs can play in a reasonable manner. The Game Strategies introduce external knowledge of gameplay tactics, enhancing the LLM's performance and simulating a human-like playstyle. Prompting Techniques involve the methods used to guide the LLM, such as Chain-of-Thought (CoT). This part also dictates how the LLM utilizes the game knowledge, for example, by referencing specific game rules or following customized strategies during decision-making.
    
\end{itemize}

\subsection{Choice of Game}

In the following, we outline the principles used to select games for LLMs to evaluate, considering both the game design and the capabilities of LLMs.

First, \textbf{all necessary game information must be representable as text}. This follows standard practice in existing LLM-based game agent research~\cite{ma2023large}. Consequently, games that rely heavily on visual information, such as First-Person Shooters, are excluded from our current selection.  For this study, we focus on games that can be expressed in text form (though not necessarily text-based games) and evaluate them using LLMs like GPT-3.5 and GPT-4.



Second, we aim to test our framework on \textbf{the original versions of real, widely played games}.
Popular games often achieve their status through polished and balanced design, making them engaging and successful across diverse audiences. Thus, these games are more representative of real-world scenarios encountered in game development. By focusing on popular, full-scale games, this evaluation seeks to provide more applicable and relevant insights for future research in LLM-based game testing.


Third, to evaluate how closely LLMs can represent game difficulty as experienced by human players, we need to choose games that \textbf{have publicly available human gameplay data}. Comparing LLMs' performance with human gameplay data is the most straightforward and convincing method for determining how accurately LLMs assess game difficulty relative to real players. 
Using games with available human play data establishes a shared foundation for LLMs and humans, allowing difficulty assessments to be made under the same conditions.

Based on these considerations, we selected two games for our evaluation: \textit{Wordle} and \textit{Slay the Spire}. These widely regarded, award-winning games represent high-quality game design and vary significantly in the complexity of their rules and action space. More information on the game rules and how they align with our criteria will be provided in the following sections.

\section{Wordle}

\subsection{Background}
\textit{Wordle} is a web-based word puzzle game published by The New York Times\footnote{https://www.nytimes.com/games/wordle/index.html} which has approximately 300,000 active daily users.
In this game, players have six attempts to guess a five-letter word. After each guess, players receive feedback indicating how close their guess is to the correct word, and their goal is to arrive at the correct answer in as few guesses as possible. 

More specifically, when a player makes a guess, the game provides feedback by highlighting each letter in one of three colors: green, yellow, or gray. A green letter means the letter is in the correct position in the word. A yellow letter means that the letter is in the word but in a different position. A gray letter indicates that the letter is not in the word at all. For example, if the target word is ``APPLE'' and the player guesses ``ALERT,'' the ``A'' would be highlighted in green, the ``L'' and ``E'' in yellow, and the other letters in gray. Players use this feedback to adjust their subsequent guesses, aiming to identify the correct word within the six attempts. 

As a game focused purely on wordplay, \textit{Wordle} is an ideal example for our use case because the entire game environment can be described in text. Solving the puzzle requires strategic reasoning, as players must balance the trade-off between using untested letters to gather more information or relying on tested letters to make a strategic bet on the next guess. Additionally, the puzzles are independent of each other, which allows us to assess the difficulty level of each one individually.

Furthermore, \textit{Wordle} offers publicly available human play data. Hosted on The New York Times website, the game has an accompany tool of \textit{WordleBot}, which records and analyzes player performances. \textit{WordleBot} documents over 600,000 plays from real users, providing detailed statistics such as the average number of attempts required to solve the puzzle and the distribution of scores. This dataset~\cite{EngagingData2024} allows us to assess how closely the LLM's assessment of puzzle difficulty aligns with that indicated by human performance. Due to \textit{Wordle}'s simple game mechanics, the difficulty of each puzzle can be directly measured by the average number of attempts needed to solve it or by the percentage of players who solve the puzzle within six guesses (commonly referred to as the win rate). Generally, the more attempts required or the lower the win rate, the more challenging the puzzle is considered.

\subsection{Implementation}
\paragraph{Game Engine and I/O}
Thanks to \textit{Wordle}'s straightforward game rule, we were able to implement the game locally on our machine without the need to connect to the web server. The core process of \textit{Wordle} is essentially verifying whether the letters in the player's guess match those in the target word and providing feedback based on their position. We developed a Python program to implement this game logic. The program takes a player's guess as input and compares it to the target word. It then generates feedback by identifying which letters are correct and in the correct position, which letters are correct but in the wrong position, and which letters are not in the word at all. This feedback is described in natural language, along with additional information such as the number of guesses taken and the guess history, to be processed by the AI agent.

\paragraph{LLM Agent and Prompting}

To set up the context for the LLM to play the \textit{Wordle} game, we first prepare a generic description of the game's rules and an explanation of the input/output format from the game environment. We then provide real-time in-game information as the feedback of the last attempt, and instruct the LLM to generate the next guess.

\begin{figure*}[t!]
    \centering
    \includegraphics[width=1.0\textwidth]{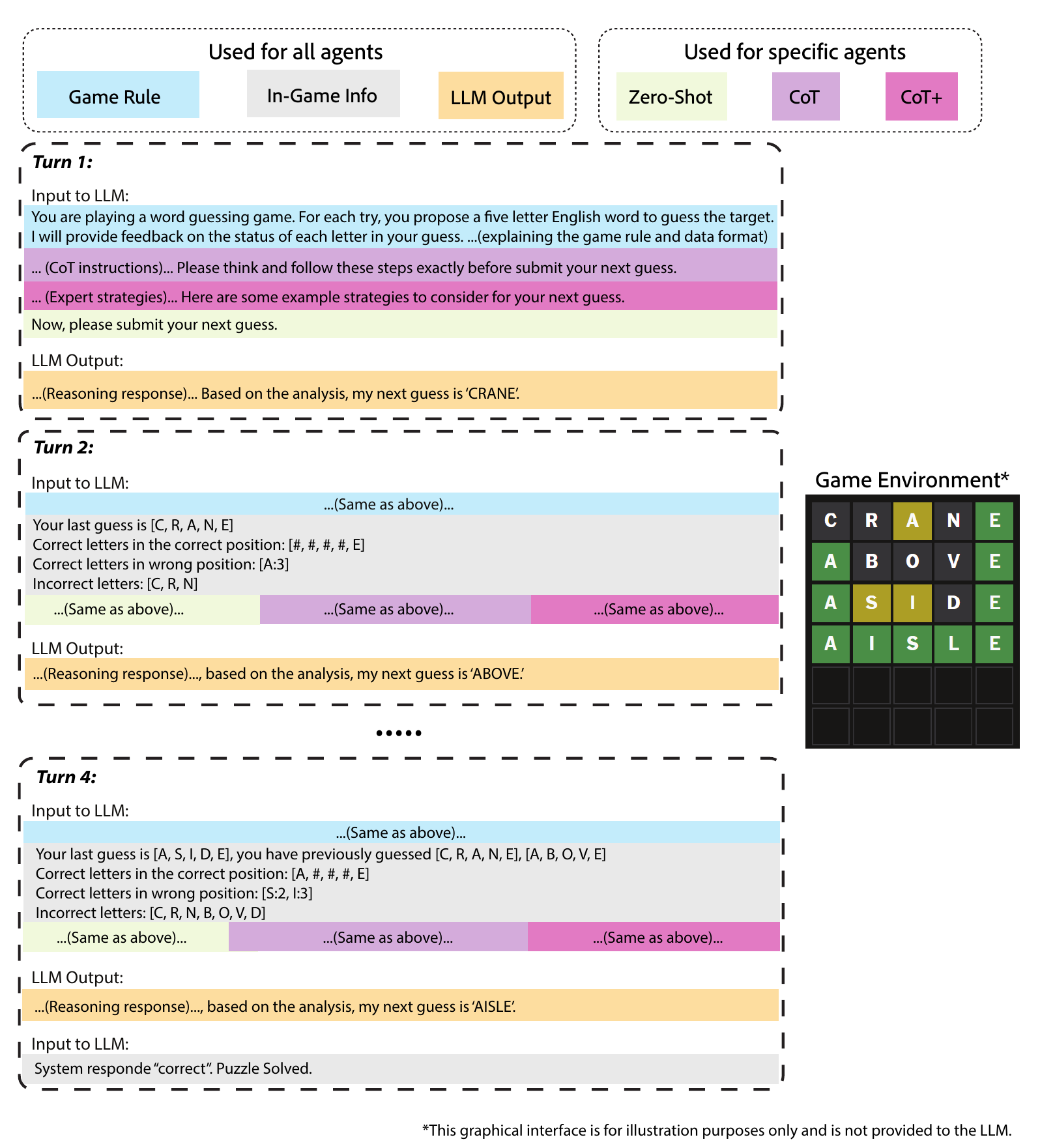}
    \caption{An example run of the LLM agent solving a \textit{Wordle} puzzle. The LLM is initially provided with the game rules and type-specific prompt (e.g., Zero-Shot, CoT and CoT+). The LLM then generates its first guess and receives feedback from the game regarding correct and incorrect letters and their positions. Using this feedback, along with the same prompt from the previous turn, the LLM produces its next guess. This cycle continues until the LLM either guesses the correct word or exceeds the maximum number of allowed guesses.}
    \label{fig:wordle}
\end{figure*}

To evaluate how different prompt engineering techniques affect the LLM agent's ability to play and how closely this ability mirrors human performance, we use three distinct prompting methods. The first method is \textbf{zero-shot prompting}, where we simply describe the game rules and ask the LLM to generate a guess directly after receiving in-game information. The second method is \textbf{CoT} prompting. In this approach, we not only provide the game rules but also prompt the LLM to engage in step-by-step reasoning before making a guess. The third method combines \textbf{CoT with game strategies}, denoted as CoT+. Given that \textit{Wordle} is a widely popular game with extensive online discussions, we incorporate expert strategies for playing \textit{Wordle} found online and guide the LLM to reason through these strategies using CoT. This approach is inspired by existing work on developing effective agents for strategy games~\cite{huang2024pokergpt, saha2024language, hu2024pok}, and is expected to enhance the LLM's gameplay ability. However, we do not fine-tune the LLM to incorporate this knowledge, as our focus is on utilizing off-the-shelf LLMs for general gameplay. See \figref{wordle} for an illustration of an example round where our LLM agent plays the Wordle game, along with sample interactions between the LLM and the game engine.

We use two different LLM models, GPT-3.5 Turbo and GPT-4, resulting in a total of six different configurations.

\paragraph{Baselines}
For comparison, we use an open-source \textit{Wordle Solver} agent\footnote{https://github.com/jason-chao/wordle-solver}, based on information theory, designed to minimize the number of guesses required to solve the game.  The \textit{Wordle Solver} works by using the feedback from each guess to eliminate words that do not compatible with current condition, gradually narrowing down the list of possible words. After the first two or three guesses, the list of potential words is often still quite large. At this point, the solver calculates which letters are most likely to appear in the correct word and suggests guesses based on those high-probability letters. The \textit{Wordle Solver} is considered a near-optimal player, achieving an average of 3.55 steps per puzzle with a 100\% win rate, outperforming the global human average of 3.97 steps~\cite{EngagingData2024}.
Comparing this baseline with other LLM agents will help us understand if better gameplay performance translates into more accurate simulation of human players.

\subsection{Experiment}
The goal of this study is to explore two key questions: \textbf{(i) Are puzzles that are difficult for human players also challenging for LLM agents, requiring more steps to solve?} Even if some LLM agents may generally take more guesses than humans, we are interested in whether they follow a similar trend in terms of relative difficulty. In other words, we want to determine if the LLM agents' performance correlates with human on which puzzles are more challenging or easier, focusing on the relative difficulty among different puzzles rather than the absolute number of guesses. \textbf{(ii) Do the LLM model and prompting methods influence the correlation between LLM performance and human play? If so, how?} Understanding this is crucial for identifying which LLM models and prompting techniques align more closely with the human play experience.

To address the key questions, we collected \textit{Wordle} puzzles from the dataset~\cite{EngagingData2024} spanning the period from March 7, 2024, to August 16, 2024, resulting in a set of 529 distinct puzzles. 

To evaluate each agent's performance, we had them attempt to solve every puzzle.
Due to the inherent randomness in the LLM agents, we conducted 20 trials per puzzle and calculated the average number of guesses required to solve each one. For the \textit{Wordle Solver} agent, which uses a deterministic algorithm, we ran the simulation only once per puzzle. We observed that under the original rule of a maximum of six guesses per puzzle, some LLM agents may fail to complete the task at most of the time. To obtain more diverse statistics for analysis and reduce the number of failure cases, we increased the cap of allowed guesses to 12, providing greater granularity in assessing each agent's performance.

We used the average number of guesses for each puzzle as an indicator of difficulty, as a higher number of guesses intuitively suggests a more difficult puzzle. Each agent reported the number of guesses for each of the 529 puzzles. After obtaining the results for all puzzles from each agent, we conducted the Pearson correlation test between the number of agent’s guesses and the corresponding average guesses made by human players. 


\paragraph{Results}

The results of our experiment are summarized in \tabref{wordle_results}. The first row presents the aggregated average number of guesses used by each agent across all puzzles. This can be regarded as a key indicator of their overall problem-solving ability. Following this, the Pearson correlation coefficient ($r$) and the $p$-value from the significance test are shown for each agent. As illustrated by the table, the Wordle Solver demonstrates the strongest \textit{Wordle} gameplay ability by requiring the fewest guesses on average, even outperforming humans. In contrast, LLM agents display an average number of guesses ranging from $5.12$ to $9.35$, which is worse than the human average. GPT-4 generally uses fewer guesses than GPT-3.5 with the same prompting techniques. Notably, the implementation of advanced prompting techniques such as CoT reasoning and the integration of game strategies largely enhance the LLM agents' gameplay performance.

Despite performing below human levels, the average number of guesses by GPT-4 CoT/CoT+ agents shows a moderate to strong correlation with the human average, with all correlations being statistically significant ($p < 0.001$). This finding suggests that puzzles hard for human players are similarly difficult for LLM agents, which indicate that \textbf{the number of guesses used by LLMs could serve as a reliable proxy for puzzle difficulty in \textit{Wordle}}. On the other hand, the Wordle Solver, despite its near-optimal gameplay, exhibits a very weak correlation with human performance, and this correlation is not statistically significant ($p > 0.05$). We attribute this to its algorithmic nature, which optimizes information entropy reduction with each guess, a strategy not typically feasible by human players. This discrepancy suggests the limitation of using a highly optimized AI for game difficulty testing, as it does not necessarily reflect human problem-solving approaches.

Within the LLM agents, further insights were observed. Overall, as the gameplay performance of LLM agents improves, their correlation with human performance also increases. For instance, the zero-shot agents of both GPT-3.5 and GPT-4 exhibit low gameplay ability and weak correlation with human players, likely because the models lack an understanding of basic Wordle-solving tactics, leading to behavior similar to random guessing. However, with simple prompting, both gameplay performance and human correlation improve largely. This suggests that while a zero-shot LLM may not be an effective player or tester, its performance and alignment with human can be enhanced through prompting. Additionally, more advanced LLMs, such as GPT-4, perform better in terms of both gameplay ability and human correlation when CoT reasoning and external strategies are applied. This can be attributed to GPT-4's superior ability to follow instructions and reason accurately. These findings suggest that \textbf{when testing game difficulty, more powerful LLMs, coupled with prompting techniques like CoT and strategic knowledge, should be used to simulate human play more closely.}

In conclusion, these results highlight the potential of GPT agents to serve as proxies for human players in assessing the relative difficulty of a set of \textit{Wordle} puzzles, especially when enhanced with reasoning and strategic knowledge.



\begin{table*}[t!]
\centering
\begin{tabular}{c|ccccccc|c} 
 \toprule
 \multirow{2}{*}{Agent}  & \multirow{2}{*}{Wordle Solver} & 
 \multicolumn{3}{c}{GPT-3.5}  &  \multicolumn{3}{c|}{GPT-4} & \multirow{2}{*}{Human}  \\ 
  & & ZS & CoT & CoT+ & ZS & CoT & CoT+ &  \\
 \hline
 Avg. Guesses & 3.55 & 9.35 & 7.65 & 7.41  & 7.79 & 6.802 & 5.12 & 3.97 \\ 
 $r$          & \cellcolor{gray!10}.075 & \cellcolor{gray!35}.237 &  \cellcolor{gray!35}.365 & \cellcolor{gray!35}.387  & \cellcolor{gray!35}.259 & \cellcolor{gray!55}.435 &  \cellcolor{gray!75}.624 & - \\
 $p$          & .124 & <.001 & <.001 & <.001  & <.001 & <.001 & <.001 & - \\

 \bottomrule
\end{tabular}
\caption{\textit{Wordle} results for different agents. ZS indicates Zero-Shot prompting, CoT indicates the use of Chain-of-Thought reasoning, and CoT+ indicates Chain-of-Thought with external strategy information. We use gray boxes, ranging from lighter to darker, to indicate no or very weak correlation ($0 < r < 0.2$), weak correlation ($0.2 < r < 0.4$), moderate correlation ($0.4 < r < 0.6$), and strong correlation ($r > 0.6$).}
\label{tab:wordle_results}
\end{table*}

\section{Slay the Spire}

\begin{figure*}[t!]
    \centering
    \includegraphics[width=0.85\textwidth]{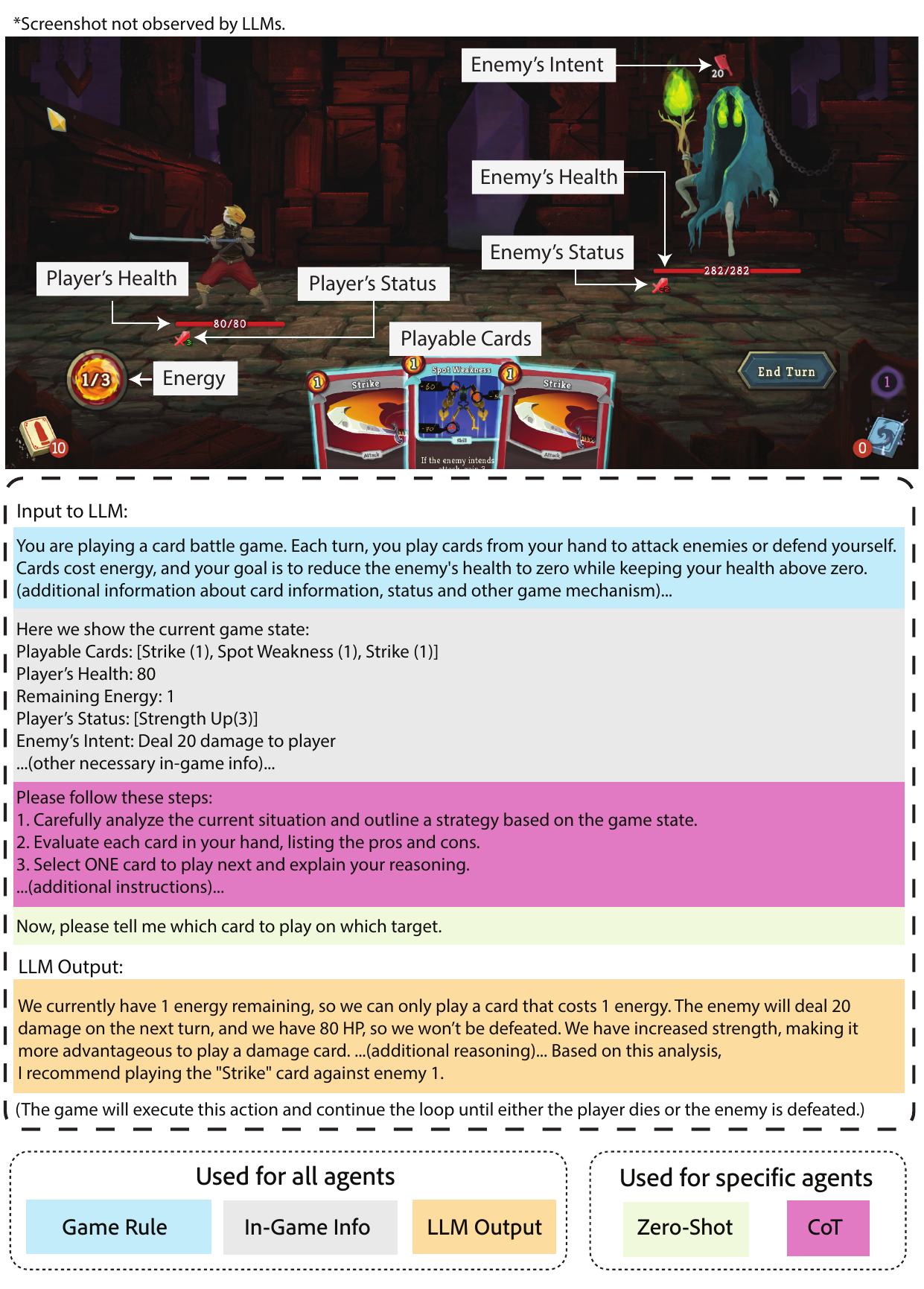}
    \vspace{-3mm}
    \caption{An example of a turn in \textit{Slay the Spire} played by LLMs. The top figure shows a screenshot of the game, along with information that can be perceived by a player. Below is an example interaction between the game and the LLM. }
    \label{fig:sts}
\end{figure*}

\subsection{Background}
\textit{Slay the Spire} (StS)~\cite{slay_the_spire} is an award-winning single-player video game that combines deck-building with roguelike~\footnote{Roguelike refers to a subgenre of video games characterized by procedurally generated levels, high randomness, and permanent character death, encouraging repeat playthroughs where each run offers a unique experience.} elements. It features a card battle system known for its intricate mechanics, which demand strategic planning and effective card synergy.
In the gameplay, players progress through levels, using their evolving deck to battle random enemies. After each victory, the player selects a new card from a random set to add to their deck, gradually strengthening the deck to face increasingly challenging enemies. 

StS could be a representative candidate for testing by LLM agents for several reasons. First, all game mechanics, including the rules, card descriptions, and enemy strategies, can be represented as text. There is no need to use method for graphics understanding to execute gameplay. 

Second, the game requires continuous understanding of the current game state, such as the cards in hand and the enemy's health points. Players must constantly consider optimal strategies and calculate odds to maximize their chances of winning. To illustrate why StS is an appropriate environment for our experiments, consider the following example cards: 
\begin{itemize} \item Strike (cost 1): Effect: Deal 6 damage. \item Bash (cost 2): Effect: Deal 8 damage. Apply \texttt{vulnerable} to the enemy. \end{itemize}
As shown above, the Strike card deals a straightforward 6 damage, but if a card like Bash is used first, applying \texttt{vulnerable}, the enemy will take 50\% more damage from subsequent attacks. Therefore, when Strike is played on a \texttt{vulnerable} enemy, it deals 9 damage instead of 6. This highlights the importance of card play order during combat. With more than 100 different types of cards, there is a wide range of possible combinations that LLMs can reason about.

Third, the developers have released a dataset of real human gameplay, containing 75 million+ runs of game sessions\footnote{https://foxrow.com/slay-the-spire-statistical-analysis}. This dataset enables us to analyze the human performance when combating different enemies and compare it with our LLM agent or other baselines.

It is also worth noting that StS has been used in previous work involving LLMs in gameplay~\cite{bateni2024sts}. However, their focus is primarily on developing LLM agents using various prompting techniques to play a simplified version of StS and analyze its performance, without comparing it to real human data. In contrast, we aim to evaluate the \textbf{original} version of StS using a generic LLM framework, with the goal of understanding the feasibility of using LLMs to measure game difficulty.

\subsection{Implementation}

The original StS is a closed-source game released on video game platforms like \textit{Steam} and \textit{Xbox}, which do not support a programming interface. However, thanks to the \textit{Steam Workshop}~\footnote{https://steamcommunity.com/workshop/} community, many developers have built \textit{Mods} \footnote{extensions that can be installed with the original game to modify its mechanics or provide support tools during gameplay} that can be utilized for automated gameplay. For our experiment purposes, we installed two Mods to implement the automation system.

The first is \textit{BasicMod}, a fundamental Mod required for the operation of other Mods. It provides the infrastructure that allows additional Mods to run on top of the original game, along with a console feature that enables players to use command lines to directly modify game elements, such as the player's health, the enemies encountered, or the cards in hand.

The second Mod is \textit{CommunicationMod}, which does not modify gameplay but enables the extraction of game information and submission of input actions via standard streams (\textit{stdin/stdout}). It also offers a Python interface, allowing interaction with the game through APIs.

With these two Mods, we can easily treat the game environment as a black box and allow LLM to interact with it through the API. The Python interface serves as the game I/O component, converting all game information into text and translating the LLM's output into executable game actions.

\paragraph{LLM Agent and Prompting}
In this study, we use two types of LLM agent: \textbf{Zero-Shot} Agent with game rules, and \textbf{CoT} Agent with both game rules and Chain-of-Thought reasoning.

To prompt the Zero-Shot LLM agent, we combine game rules with real-time game information obtained through the \textit{CommunicationMod} module, forming a natural language input that includes the basic objectives, core mechanics, cards in the deck, along with player and enemies' statuses (e.g., hp, block, and \texttt{vulnerable} as mentioned in the previous example). Based on this input, the LLM is asked to generate the next card to play and the target of the card. 
Once the LLM determines the next action, the response will be parsed by a Python script and sent to the \textit{CommunicationMod} module as executable action. The action will then be executed by the game environment as if they were human inputs, allowing the game to run normally. After the player ends their turn, the enemy will take its actions, and the LLM will receive updated in-game information to repeat the process. This loop will continue until the player's health reaches zero (resulting in failure) or all enemies are defeated (resulting in a win).

In addition to the basic game knowledge mentioned above, the CoT agent is provided with step-by-step reasoning instructions. In CoT prompting, we ask the LLM to analyze the in-game situation, determine the optimal strategy, and evaluate each playable card before selecting the card to play. This process encourages the agent to exhibit more logical and human-like behavior.
To keep the prompt concise, only high-level instructions are provided, while avoiding detailed tactics on specific cards or combinations.
This approach also allows us to assess how much strategy LLMs can generate independently and how well they can simulate human play when no external strategy is given.

As in the \textit{Wordle} experiment, we use both GPT-3.5 Turbo and GPT-4 for each agent, resulting in a total of four different LLM configurations. An example turn of LLM agents playing StS is shown in \figref{sts}.

\paragraph{Baseline Agents}
Similar to the \textit{Wordle} experiment, to evaluate how well LLMs can approximate human behavior in StS, we compare their performance with two baseline agents, both of which follow rule-based behavior. The first baseline is a random behavior agent, which randomly selects and plays card from the available cards, ending the turn when there is not enough energy to play another card or when all cards in hand have been played.

The second baseline is an open-source AI\footnote{https://github.com/ForgottenArbiter/spirecomm} for playing StS, which incorporates expert knowledge on tactics and card priorities, and follows a hard-coded gameplay logic. Details of the method can be found in the source code. According to its documentation, this agent is capable of achieving a 80\% win rate in defeating all bosses.

These two baselines allow us to assess how closely the difficulty indicated by LLM performance aligns with that of human players, in comparison to heuristic AI. In other words, we aim to determine whether LLMs can provide a more accurate measure of game difficulty than heuristic-based AI.

\subsection{Experiment}

We tested the agents by having them battle all the bosses they could encounter in Act 1 and Act 2 (``Act'' is a game-specific term equivalent to ``Level''). To maintain a controlled experiment, we fixed the character as \textit{Ironclad} and used a well-balanced, representative deck. We found that when using a deck of average strength, similar to that of a human player in the corresponding act, the LLMs had a win rate of less than 5\% against the bosses. Using a stronger deck helped compensate for the LLM's sub-human gameplay performance and ensured an average win rate of 65\%.



As an indicator of human-perceived enemy difficulty, we use the average win rate of each boss, calculated from gameplay data released by the developers. This metric provides a baseline for how challenging human players find each boss, offering a reference point for comparison with agent performance.

To assess the agents' gameplay performance, we use ``HP remaining'' as an indicator of enemy difficulty. This metric, which reflects the agent's remaining health at the end of each battle, provides more nuanced insights into difficulty than win rate alone, as it indicates not only whether the agent won but also how easily it did so. In other words, higher remaining HP suggests a less challenging battle. We did not use HP as a performance indicator for human players because, due to certain game mechanics, human players may start boss battles with varying amounts of initial HP. This inconsistency would make comparisons across human play unreliable.

After completing 20 trials with each agent against every boss, we conducted a Pearson correlation test. We examined the relationship between the agent's remaining HP and the human-indicated difficulty derived from the average win rate by human player. If the agent was defeated, we recorded a remaining HP of 0, ensuring consistency in the data and allowing us to evaluate how closely the agents' difficulty perceptions align with those of human players. Ideally, since both a higher win rate and higher HP remaining indicate lower difficulty, these two metrics should show a strong positive correlation.

\paragraph{Quantitative Results}

\begin{table*}[t!]
\centering
\begin{tabular}{c|c|cccccccc} 
 \toprule
  \multirow{2}{*}{Act} & \multirow{2}{*}{Agent}  & \multirow{2}{*}{Random} & 
 \multicolumn{2}{c}{GPT-3.5}  &  \multicolumn{2}{c}{GPT-4} & \multirow{2}{*}{Rule-based Expert AI}  \\ 
  & & & ZS & CoT & ZS & CoT & \\
 \hline
 \multirow{3}{*}{Act 1} & Avg. HP & 12.717 & 8.650 & 19.917 & 22.550 & 36.733 & 32.400 \\ 
 & $r$ &  \cellcolor{gray!35}0.231 &  \cellcolor{gray!55}0.471 &  \cellcolor{gray!55}0.513  &  \cellcolor{gray!75}0.657 &  \cellcolor{gray!75}\textbf{0.871} &  \cellcolor{gray!75}0.742 \\
 & $p$ & 0.075 & <0.001  & <0.001  & <0.001 & <0.001 & <0.001 \\
  \hline
 \multirow{3}{*}{Act 2} & Avg. HP & 2.733 & 0.267 & 3.617 & 4.250 & 17.583 & 16.133  \\ 
 & $r$ & \cellcolor{gray!35} 0.287 &  \cellcolor{gray!35}0.208 &   \cellcolor{gray!35}0.395 & \cellcolor{gray!55} 0.479 &  \cellcolor{gray!75}\textbf{0.710}  &  \cellcolor{gray!55} 0.482 \\
 & $p$ &0.026  & 0.111 &  0.002 & <0.001 & <0.001 & <0.001 \\
 \bottomrule
\end{tabular}
\caption{StS results for different agents.}
\label{tab:sts_results}
\end{table*}

The results of the correlation tests are presented in \tabref{sts_results}. Similar to our findings in the \textit{Wordle} experiment, the results provide positive answers to the two key questions we posed.

First, enemies that are difficult for human players are also hard for LLM agents. This is evidenced by the very strong correlation---up to 0.871---between GPT-4 with CoT and the human-indicated difficulty derived from human gameplay.

Second, the influence of the LLM model and prompting method on the correlation mirrors that observed in the \textit{Wordle} experiment. For both Act 1 and Act 2, GPT-4 CoT shows the strongest correlation among all agents, followed by the rule-based expert AI, GPT-4 ZS, and GPT-3.5 CoT. The least capable agent is GPT-3.5 ZS, which performs and correlates even worse than a random agent in Act 2. This indicates that playing complex and strategic games like StS requires state-of-the-art models and advanced prompting techniques.

Additional observations further support these conclusions. Agents with higher average remaining HP also exhibit higher correlations with human-indicated difficulty, and this applies not only to LLM-based agents but also to the rule-based agent. This trend contrasts with our observations from the Wordle solver, which, despite its superior performance, showed poor correlation with human difficulty assessments. A possible explanation is that all agents (besides random) in the StS experiment are designed to approximate human gameplay strategies, whether they are LLMs or rule-based experts. Due to the increased complexity of StS, there is currently no agent that can outperform humans without following reasoning processes similar to those of humans, unlike the solver in \textit{Wordle} use optimization-based action. Therefore, the closer an agent's behavior pattern is to that of humans in this game, the better it can approximate enemy difficulty as indicated by human players.

Furthermore, when examining the average remaining HP, we find that the rule-based expert AI performs almost as well as the best LLM agent, GPT-4 with CoT prompting. However, its correlation with human win rates is much lower than that of GPT-4 with CoT. This indicates that even with similar gameplay capabilities, LLMs are better than heuristic AIs at reasoning and acting like humans, and thus are better at approximating human-indicated difficulty.


\paragraph{Qualitative Results}

We also present a qualitative analysis of the gameplay abilities of the GPT-4 CoT, demonstrating its human-like decision-making based on threat assessment, resource management, and optimal card usage.
\figref{sts_qualitative} shows an example of the LLM's response with step-by-step reasoning and effective decision-making.

First, when analyzing the combat, the primary threat, the \textit{Acid Slime}, is correctly identified due to its intent to attack with 16 damage. Prioritizing threats based on potential damage is a fundamental strategy in StS, ensuring the player minimizes incoming damage while managing other enemies later.

Second, the LLM performs calculations and reasoning to evaluate each card based on real-time in-game information. This is particularly important for cards such as Heavy Blade, where the final damage scales 3 times with the player's current \textit{Strength}, or \textit{Spot Weakness}, which can only be used against an attacking enemy. In addition to reasoning about the current game state, it also shows the ability to predict future game outcomes after a proposed action, correctly calculating the 59 damage of \textit{Heavy Blade} after playing \textit{Spot Weakness} to increase \textit{Strength}.

Third, given the game state and cards in hand, the LLM proposes an expert-level action plan. In this particular turn, the player has both offensive and defensive options. Playing \textit{Spot Weakness} maximizes the damage output of \textit{Heavy Blade}, ensuring that the primary threat is eliminated efficiently. By focusing on offense rather than defense, this action plan avoids the need for block cards, which would mitigate damage but not eliminate the threat.

The reasoning and strategy generated by the LLM highlights key aspects of StS gameplay: leveraging synergies between cards, anticipating future game states, and prioritizing threats based on their potential impact. We would like to emphasize that this CoT agent was not provided with any human-crafted strategic information on how cards should be played; rather, it independently developed strategies that mimic human thought processes. 

\begin{figure}
    \centering

\begin{tcolorbox}[colback=gray!5!white, colframe=gray!75!black] 
\small

\textbf{Analysis and Strategy}

Current Energy: 3 \\
Current HP: 64 \\
Current Block: 0 \\
Current Strength: 12 \\

Enemies:
\begin{itemize}
    \item Spike Slime: 52 HP, Intent: Debuff
    \item Acid Slime: 52 HP, Intent: Attack, 16 damage
\end{itemize}

Given the situation, the primary threat is \textbf{Acid Slime} due to its 16 damage attack. To minimize HP loss, we need to focus on either blocking the damage or reducing the threat by dealing damage effectively.

\textbf{Card Evaluation}

\begin{enumerate}
    \item \textbf{Defend}
    \begin{itemize}
        \item Pros: Provides 5 block for 1 energy.
        \item Cons: Doesn't contribute to offense.
        \item Overall: Basic block card; only useful if we need to mitigate impending damage right now.
    \end{itemize}

    \item \textbf{Shrug It Off}
    \begin{itemize}
        \item Pros: Provides 8 block for 1 energy and draws 1 card.
        \item Cons: Slightly better than Defend due to the card draw.
        \item Overall: Useful for both blocking and cycling through the deck.
    \end{itemize}

    \item \textbf{Heavy Blade}
    \begin{itemize}
        \item Pros: Deals significant damage due to Strength scaling. (14 + 3*12 = 50)
        \item Cons: Costs 2 energy.
        \item Overall: Excellent offensive card given current Strength.
    \end{itemize}

    \item \textbf{Spot Weakness}
    \begin{itemize}
        \item Pros: Increases Strength by 3 if an enemy intends to attack.
        \item Cons: Conditional; only useful if facing an attack intent.
        \item Overall: Useful here as Acid Slime intends to attack.
    \end{itemize}
\end{enumerate}

\textbf{Optimal Play}

The best initial play is \textbf{Spot Weakness} to further bolster our Strength to 15, making subsequent attacks even more potent. This prepares \textbf{Heavy Blade} for maximum effectiveness (14 base + 3*15 = 59 damage), which is more than enough to eliminate \textbf{Acid Slime} without taking any damage.

\textbf{Next Card to Play:} Spot Weakness

\textbf{Target:} Acid Slime

\end{tcolorbox}
\vspace{-10pt}
\captionof{figure}{An example response from GPT-4 when prompted with CoT to play StS.}
\label{fig:sts_qualitative}
\vspace{1pt}
\end{figure}

This demonstrates the LLM's capability to produce advanced decision-making strategies commonly used by human players, showcasing its potential to offer both effective gameplay and human-like reasoning.

\section{Discussion}
In this section, we share insights from our experiences with \textit{Wordle} and \textit{Slay the Spire} and synthesize key guidelines on how to effectively use LLMs to measure game difficulty.

\textbf{G1: Text representation format matters.} During our experiment, we found that the way game information is represented for LLM processing is crucial. Game-specific information should be presented in natural language and in a manner that aligns with the LLM's inherent processing structure.
For example, in the \textit{Slay the Spire} experiment, game-specific mechanics (e.g., \texttt{Exhaust} a card) should be explained in a way that is understandable to an average person without extensive gaming experience.
Moreover, in the \textit{Wordle} experiment, we initially formatted all the 5-letter words in plain text. However, LLMs process text by splitting words into common subwords. For instance, the word ``APPLE'' might be tokenized into a sequence of ``APP'' and ``LE'' rather than a sequence of five individual characters. This tokenization phenomenon limits the LLM's ability to accurately understand the position of letters within a word. To address this, we reformatted the words explicitly as lists (e.g., ``[A, P, P, L, E]''), utilizing the LLM's ability to process structured input, and observed a substantial improvement in gameplay performance.

\textbf{G2: Some games may require compensation mechanisms to accommodate LLM performance.} As observed in our experiment, in most cases, LLMs do not perform as well as average human players. Additionally, it is possible that LLM agents may consistently fail to accomplish certain tasks. For instance, they may exceed the original limit of 6 guesses in \textit{Wordle} or have very low win rates against specific bosses in \textit{Slay the Spire}. If the win rates are consistently low and no distinction between easy and difficult challenges, it becomes hard to assess relative difficulty for these challenges using LLMs. In our practice, we increased the guessing limit in \textit{Wordle} to give the LLM a better chance of success, resulting in greater variation of LLM's performance across puzzles. In \textit{Slay the Spire}, we provided the LLM with a stronger deck than an average human player would have at the same level, ensuring the LLM could consistently defeat most bosses. However, the method of compensating for LLM performance should be considered on a case-by-case basis. It is crucial not to overcompensate, as this could result in the LLM consistently completing all challenges with exceptionally high performance. Also, the compensation should not drastically change the game mechanics in a way that no longer aligns with the human gameplay experience.

\textbf{G3: Use LLMs to understand relative difficulty across various challenges and design the difficulty curve.} Similar to the observation from the previous guideline that LLMs generally do not perform as well as the average human player, they may not be suitable for assessing the difficulty of a single isolated challenge or predicting the potential win rate for human players on that specific challenge. However, relative difficulty can still provide valuable insights for game designers when shaping the difficulty curve to better align with their intended design~\cite{juul2013art}. Moreover, once human performance has been calibrated on one or a few challenges in the game, LLMs can be used to measure the relative difficulty of new challenges with respect to the calibrated ones. This allows us to project human performance on the new challenges based on their relative difficulty.


\textbf{G4: Use more advanced LLM models and prompting techniques.} Our experiment shows that larger LLM models (e.g., GPT-4 vs. GPT-3.5) consistently show better reasoning and instruction-following abilities. Therefore, when possible, developers should use more capable LLM models. Additionally, prompting techniques, such as CoT prompting, can significantly improve the LLM’s correlation with human gameplay by encouraging the model to reason and act more like a human player. Incorporating general game strategies can also improve correlation, but it is important that these strategies reflect typical gameplay techniques rather than exploitative ``hacks,'' as the latter may lead to biased or inaccurate difficulty estimates.

\textbf{G5: Small-scale human data from pilot studies can be useful in determining various factors.}
While our framework is purely LLM-based and does not require human data, experimenting with small-scale human data as a pilot study can help optimize certain parameters or configurations within the framework. For example, it can assist in identifying which difficulty indicators, such as scores, time taken, or win rates, best correlate with human performance. Additionally, human data can provide insights into the differences in gameplay performance between humans and LLMs, helping to inform decisions on how much compensation is needed for LLM performance.




\paragraph{Future Work}
As the first work aimed at understanding the feasibility of using LLM to measure game difficulty, our work addresses a defined set of questions. However, several areas remain open for further evaluation and exploration in future research.

First, our work focuses on using LLMs to solve individual, isolated challenges within a game. However, many games feature cumulative effects, where factors such as resources, health, or  character levels carry over from one challenge to the next, influencing the difficulty of subsequent challenges. As a result, the sequence in which challenges are presented can impact their collective difficulty and the shape of difficulty curve. In future work, we aim to consider these cumulative effects into our framework by adding gameplay history and simulating LLM agents across a sequence of challenges.

Second, our current framework treats the LLM as an agent with fixed abilities, responding only based on the external information provided. This does not reflect the learning process of a human player, whose skills evolve over time. In real scenarios, as players attempt challenges, they improve and learn strategies that help them tackle future challenges. We are interested in exploring this dynamic by allowing LLMs to learn and improve through multiple trials of a single challenge. This could help estimate how human players progressively develop skills and adapt to challenges. Additionally, it would be useful to simulate different play styles, such as aggressive versus conservative play, by injecting varied play strategies as prompts for the LLM.

Third, game testing involves more than just measuring difficulty. It would be worthwhile to investigate how LLM agents could evaluate other aspects of game quality, such as identifying bugs, detecting gameplay imbalances, or assessing user experience factors.

Finally, our current framework is text-only. However, games with intensive graphical elements may not fully translate into text, or important information may be lost during the conversion. In the future, as more advanced vision-language models capable of accurately understanding game scenes become available, it would be interesting to extend our framework to more visually intensive games and conduct similar evaluations.


\section{Conclusion}
In this work, we present an LLM-based framework for measuring game difficulty. Our framework is adaptable to a wide range of games, and we have specifically tested it on \textit{Wordle} and \textit{Slay the Spire}. Our experiments demonstrate that LLMs exhibit a high correlation with human players in difficulty assessment, requiring only simple and generic prompting. Compared to rule-based agents, which take significant time to develop, and machine learning-based agents, which require extensive computational resources for training, LLMs offer a more generalized capability to play a variety of games with minimal adjustment. We hope this work opens the door for new approaches in game testing within the gaming industry, while also inspiring further applications of LLMs in this field.



\bibliographystyle{ACM-Reference-Format}
\bibliography{ref}
\end{document}